\documentclass[lettersize,journal]{IEEEtran}
\usepackage{amsmath,amssymb,amsfonts}
\usepackage{algorithmic}
\usepackage{algorithm}
\usepackage{array}
\usepackage[caption=false,font=normalsize,labelfont=sf,textfont=sf]{subfig}
\usepackage{textcomp}
\usepackage{stfloats}
\usepackage{url}
\usepackage{verbatim}
\usepackage{graphicx}
\usepackage{subcaption}
\usepackage{cite}
\usepackage{textcomp}
\usepackage{xcolor}
\usepackage{booktabs}
\usepackage{multirow}
\usepackage{pifont}
\usepackage{hyperref}
\usepackage{array}
\begin{document}


\title{Robust Industrial Cyber-Physical Classification Using Neuromorphic Temporal Embeddings and Hybrid SNN–XGBoost Under Machine Unlearning Attacks}


\author{Ammar Kamoona~\href{https://orcid.org/0000-0002-7441-9344}{\textsuperscript{\includegraphics[scale=0.06]{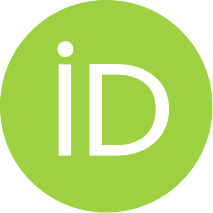}}},
Sajad Koushkbaghi\href{https://orcid.org/0000-0001-9655-5809}{\textsuperscript{\includegraphics[scale=0.06]{orcid.pdf}}},
Mahdi Jalili~\href{https://orcid.org/0000-0002-0517-9420}{\includegraphics[scale=0.06]{orcid.pdf}},~\IEEEmembership{Member,~IEEE},
Peter McTaggart~\href{https://orcid.org/0000-0002-9593-8549}{\includegraphics[scale=0.06]{orcid.pdf}},
and Xinghuo Yu~\href{https://orcid.org/0000-0001-8093-9787}{\includegraphics[scale=0.06]{orcid.pdf}},~\IEEEmembership{Senior Member,~IEEE}%
\thanks{Ammar Kamoona, Sajad Koushkbaghi, Mahdi Jalili, and Xinghuo Yu are with the EV Living Lab, RMIT University, Melbourne, VIC, Australia (e-mail: ammar.kamoona@rmit.edu.au; sajad.koushkbaghi@rmit.edu.au; mahdi.jalili@rmit.edu.au; xinghuo.yu@rmit.edu.au). Peter Mctaggart is with Powercor, Melbourne, VIC, Australia.}%
}


\markboth{Journal of \LaTeX\ Class Files,~Vol.~14, No.~8, August~2021}%
{Shell \MakeLowercase{\textit{et al.}}: A Sample Article Using IEEEtran.cls for IEEE Journals}


\maketitle

\begin{abstract}
The digitalisation of electrical distribution networks has increased the exposure of power-grid infrastructure to cyber attacks. Existing intrusion detection systems (IDSs), however, often rely on computationally expensive deep learning models that are difficult to deploy at the edge. Periodic retraining also exposes these systems to machine unlearning attacks, where selective data removal can degrade detection performance. We propose a hybrid Spiking Neural Network (SNN) and XGBoost architecture that combines efficient temporal encoding with a lightweight classifier and provides structural resilience to such attacks. The SNN is trained once on clean data and used as a fixed feature extractor, while only the XGBoost classifier is retrained during model updates. Evaluated on two real-world public power-system datasets, the proposed method achieves 99.9\% accuracy (F1-macro 0.999) on the Synchrophasor dataset and 95.0\% accuracy (F1-macro 0.943) on the MSU/ORNL dataset, outperforming standalone baselines. Under selective label-flipping attacks, the hybrid model loses only 0.9\% F1-macro at 10\% poisoning and delays target-class collapse from 60\% to 70\% poisoning compared with raw models. These results demonstrate that neuromorphic temporal encoding can provide both accurate cyber-attack detection and improved resilience to data poisoning in cyber-physical systems.
\end{abstract}

\begin{IEEEkeywords}
Spiking neural networks, intrusion detection systems, power grid cybersecurity, machine unlearning, false data injection, neuromorphic computing, surrogate gradient learning
\end{IEEEkeywords}

\section{Introduction}
\IEEEPARstart{E}{lectrical} distribution networks are increasingly evolving into interconnected cyber-physical systems as distributed energy resources and advanced control technologies become more widely integrated, resulting in reliance on intelligent electronic devices (IEDs) such as remote terminal units (RTUs) and phasor measurement units (PMUs), and communication networks for monitoring, protection, automation, and control. \cite{khalaf2024survey}.
While these technologies enable more flexible and efficient operation, they also increase the dependence of distribution networks on digital infrastructure and communication networks.
Consequently, vulnerabilities in cyber components can affect the operation of physical grid assets, making cybersecurity an integral component of the reliability and resilience of modern distribution networks \cite{liang2017Ukraine}.

Cyber attacks against power-system cyber-physical infrastructure can target both the information used to monitor and control the physical system and the communication mechanisms through which this information is exchanged.
False data injection (FDI) attacks alter measurement data to provide misleading information about the state of the power system, potentially compromising monitoring and state-estimation functions \cite{tian2022joint}.
In synchrophasor-based monitoring, time-synchronisation attacks (TSAs), particularly those based on GPS spoofing, can manipulate the timing information used to synchronise measurements, potentially compromising monitoring and control functions \cite{zhang2020review}.
At the communication layer, attacks can exploit power-system communication protocols to inject, replay, or manipulate messages, potentially disrupting the exchange of information between IEDs.
Such attacks are particularly relevant to IEC 61850-based substations, where GOOSE messages support time-critical communication for protection and automation functions \cite{quincozes2024ereno}.

Since cyber attacks can have a widely diverse nature and manifest across various aspects of a cyber-physical power system, their accurate detection is a challenging task.
Furthermore, an effective intrusion detection system should not only identify whether an attack has occurred, but also classify its type, providing more informative awareness for subsequent response and future mitigation.
When it comes to the practical implementation of a detection system, minimal computational overhead and low latency are important, particularly when monitoring must be performed continuously and close to the physical infrastructure.  
Moreover, practical detection systems may need to accommodate heterogeneous data sources, ranging from continuous-valued electrical measurements to structured communication traffic generated by grid protocols \cite{kholidy2019vhdra}.

Existing cyberattack detection approaches for power systems broadly include rule- and signature-based methods \cite{mitchell2013behavior}, model-based methods that exploit knowledge of system dynamics and operating constraints \cite{sridhar2014model}, and data-driven methods that learn attack characteristics from observed data \cite{boyaci2021gnn, haghshenas2022temporal, boyaci2021chebyshev, boyaci2021joint, niu2019dynamic}.
Rule- and signature-based methods rely on predefined thresholds, rules, or known attack signatures and therefore require the relevant attack behaviors to be specified in advance \cite{nie2024intrusion}.
Model-based approaches can exploit physical knowledge of the grid and provide interpretable detection decisions, but their effectiveness can depend on the availability and fidelity of the underlying system model and can be affected by modeling uncertainties and changing operating conditions \cite{xu2021transfer}.
Data-driven approaches provide an alternative by learning attack-related patterns from measurements or communication data, thereby reducing reliance on manually specified attack signatures and detailed physical models.
Deep learning methods have consequently been investigated for detecting cyberattacks from complex power-system measurements and communication data, with architectures such as convolutional, recurrent, and graph neural networks \cite{he2017, yin2022powerfdn, liu2024graphcci, oinonen2026cyberattack}.
However, the computational and memory requirements of conventional deep learning models can impose constraints on low-latency and resource-efficient deployment.

Spiking neural networks (SNNs) offer an alternative neural computing approach in which information is represented through discrete spike events and processed according to the temporal dynamics of spiking neurons \cite{eshraghian2023snntorch,neftci2019surrogate}. 
Their event-driven operation and sparse activity offer the potential for computationally and energy-efficient processing, while their intrinsic temporal dynamics make them suitable for analysing sequential signals. 
These characteristics have motivated the application of SNNs to intrusion detection and power-system security, including dynamic FDI attack detection and network intrusion detection \cite{hamedani2019reservoir,zhou2020single,hood2024snnpg,patel2026evaluation,mia2025neuromorphic}. 
However, existing SNN-based approaches have primarily focused on detection performance, network architecture, or energy efficiency, while the resilience of their learned representations to training-time data manipulation, such as malicious machine unlearning, remains largely unexplored. 
Training-time poisoning attacks can manipulate the learned decision function by corrupting training samples or labels \cite{paphitis2026unlearning}, highlighting the need to consider the resilience of the detection systems' training data pipelines in addition to their ability to detect cyberattacks.

To address this vulnerability, we propose a hybrid SNN--XGBoost architecture that separates representation learning from classifier adaptation. 
A Direct-SNN with Leaky Integrate-and-Fire (LIF) neurons processes sliding windows of PMU measurements and generates temporal membrane-potential embeddings, while a lightweight XGBoost classifier is periodically retrained on these fixed embeddings to accommodate evolving operating conditions. 
The SNN is trained once using verified clean data during system commissioning and subsequently frozen, preventing label manipulation during classifier retraining from altering the learned feature representation. 
Because the membrane-potential embeddings capture temporal characteristics of the underlying electrical signals rather than classifier labels, poisoned samples remain constrained by the physical structure encoded by the frozen feature extractor.

\textcolor{black}{The main contributions of this paper are as follows:}
\begin{enumerate}
    \item We show that the proposed hybrid SNN+XGBoost architecture achieves state-of-the-art detection accuracy on two real power grid datasets: 99.9\% (F1-macro 0.999) on the Synchrophasor dataset and 95.0\% (F1-macro 0.943) on the MSU/ORNL dataset, outperforming all standalone baselines.
    \item We demonstrate that the proposed method provides inherent robustness against machine unlearning attacks. At 10\% label poisoning, the hybrid loses only 0.9\% F1-macro and delays target-class collapse from 60\% to 70\% poisoning compared to raw feature models.
    \item We provide a mechanistic explanation: SNN membrane potential embeddings encode physical signal structure (i.e., the temporal dynamics of voltage and current measurements) that is invariant to label corruption.
    \item We propose a secure deployment architecture that separates the frozen feature extractor from the periodically updated classifier, providing defence-in-depth against training data poisoning.
    \item We present experiments on the MSU/ORNL dataset (78,377 samples, 128 PMU features) comparing the hybrid against raw XGBoost and Random Forest at 10 attack levels (0--90\%).
\end{enumerate}

The rest of this paper is organised as follows. Section~\ref{sec:related} reviews related work. Section~\ref{sec:preliminaries} presents the theoretical foundations. Section~\ref{sec:proposed_method} describes the proposed method. Section~\ref{sec:experiments_setup} provides experimental setup and Section\ref{sec:results} provides the results and the discussion. Section~\ref{sec:conclusion} concludes the paper.

\section{Related Work}
\label{sec:related}
\subsection{Machine Unlearning Attacks on Smart Grid IDS}
Machine unlearning was originally studied for privacy-preserving data deletion \cite{bourtoule2021machine,xu2024machine}, where the goal is to remove the influence of specific training samples from a deployed model. Paphitis et al. \cite{paphitis2026unlearning} repurposed this concept as an attack on smart grid ML systems. In their attack model, an adversary gains access to the training data pipeline and selectively relabels a percentage of a target class. When the IDS undergoes periodic retraining, it ``unlearns'' the targeted attack signatures.

Paphitis et al. showed that even 10--20\% label corruption significantly degrades detection of the targeted class, while overall accuracy remains largely unaffected. This makes the attack difficult to detect through standard monitoring. Their evaluation used conventional classifiers (Random Forest, SVM, Neural Networks) on the IEEE 118-bus system. They concluded that no model exhibited structural resilience. In contrast, we show that neuromorphic architectures with separated feature extraction provide a qualitatively different robustness profile.

Related threats include training data poisoning \cite{biggio2012poisoning}, backdoor attacks \cite{gu2017badnets}, and model inversion \cite{fredrikson2015model}. Machine unlearning is distinct in that it targets the periodic retraining process specifically, exploiting the operational necessity of model updates.
\subsection{SNNs Network for Intrusion Detection}
SNNs process information through discrete binary spikes, offering orders-of-magnitude energy reduction on neuromorphic hardware \cite{patel2026evaluation,mia2025neuromorphic}. For power grid security, Hood et al. \cite{hood2024snnpg} proposed SNNPG for binary FDI detection on synthetic data. Hamedani et al. \cite{hamedani2019reservoir} applied reservoir computing for dynamic FDI detection. Both are limited to binary classification on simulated data with fewer than 2,000 samples.

For general network IDS, Patel et al. \cite{patel2026evaluation} evaluated 27 SNN configurations on standard datasets, finding that encoding choice dominates architecture choice. Mia et al. \cite{mia2025neuromorphic} introduced continual learning with STDP for IDS adaptation. However, no prior SNN-based IDS work has considered adversarial robustness to training data manipulation. Hybrid architectures combining neural feature extractors with tree-based classifiers have been explored for tabular data \cite{popov2019neural}.

\subsection{Adversarial Robustness of IDS}
Adversarial robustness research in IDS has focused on evasion attacks (i.e., adversarial examples at inference time) \cite{apruzzese2022modeling} and certifiable robustness \cite{cohen2019certified}. Defence mechanisms include adversarial training, ensemble methods, and input preprocessing. However, these methods target inference-time manipulation and do not address training-time data poisoning.

The concept of using structural properties of a model to provide robustness has been explored in randomised smoothing and Lipschitz-bounded networks. We extend this concept to neuromorphic architectures. The SNN membrane dynamics provide structural robustness against label poisoning by encoding physical signal properties that are independent of labels.  

\subsection{Research Gaps}
Table~\ref{tab:comprehensive_comparison} compares existing approaches across eight dimensions relevant to robust IDS under adversarial conditions. Our work is the first to combine neuromorphic feature extraction with explicit robustness evaluation against training data manipulation. The table reveals that no prior work combines neuromorphic temporal feature extraction with robustness evaluation against training data manipulation. Machine unlearning works \cite{paphitis2026unlearning,bourtoule2021machine} evaluate conventional models without exploring architectural defences. SNN-based IDS works \cite{hood2024snnpg,patel2026evaluation,mia2025neuromorphic} do not consider adversarial robustness. Power grid DL methods \cite{boyaci2021chebyshev,niu2019dynamic} use temporal features but train end-to-end, making them fully vulnerable to label corruption. Our approach uniquely provides: (1)~neuromorphic temporal features, (2)~a separated feature extractor immune to periodic retraining corruption, and (3)~explicit robustness evaluation under the Paphitis et al. unlearning attack model.

\begin{table*}[!t]
\centering
\caption{Comprehensive Comparison of Existing Approaches for Power Grid IDS Under Adversarial Conditions. \ding{51} = present, \ding{55} = absent. Our work is the only approach combining neuromorphic temporal features with explicit machine unlearning robustness evaluation.}
\label{tab:comprehensive_comparison}
\renewcommand{\arraystretch}{1.15}
\footnotesize
\begin{tabular}{p{3.0cm}cccccccc}
\toprule
\textbf{Method} & \textbf{SNN/} & \textbf{Real} & \textbf{Multi-} & \textbf{Temporal} & \textbf{Separated} & \textbf{Poisoning/} & \textbf{Power} & \textbf{Periodic} \\
 & \textbf{Neurom.} & \textbf{Data} & \textbf{Class} & \textbf{Features} & \textbf{Extractor} & \textbf{MU Eval.} & \textbf{Grid} & \textbf{Retrain} \\
\midrule
\multicolumn{9}{l}{\textit{Machine unlearning and data poisoning attacks}} \\
\midrule
Paphitis et al. \cite{paphitis2026unlearning} (2026) & \ding{55} & \ding{55} & \ding{51} & \ding{55} & \ding{55} & \ding{51} & \ding{51} & \ding{51} \\
Bourtoule et al. \cite{bourtoule2021machine} (2021) & \ding{55} & \ding{51} & \ding{51} & \ding{55} & \ding{55} & \ding{51} & \ding{55} & \ding{55} \\
Biggio et al. \cite{biggio2012poisoning} (2012) & \ding{55} & \ding{51} & \ding{55} & \ding{55} & \ding{55} & \ding{51} & \ding{55} & \ding{55} \\
\midrule
\multicolumn{9}{l}{\textit{SNN-based approaches for power grid / IDS}} \\
\midrule
SNNPG \cite{hood2024snnpg} (2024) & \ding{51} & \ding{55} & \ding{55} & \ding{55} & \ding{55} & \ding{55} & \ding{51} & \ding{55} \\
Hamedani et al. \cite{hamedani2019reservoir} (2019) & \ding{51} & \ding{55} & \ding{55} & \ding{51} & \ding{55} & \ding{55} & \ding{51} & \ding{55} \\
Patel et al. \cite{patel2026evaluation} (2026) & \ding{51} & \ding{51}$^*$ & \ding{51} & \ding{55} & \ding{55} & \ding{55} & \ding{55} & \ding{55} \\
Mia et al. \cite{mia2025neuromorphic} (2025) & \ding{51} & \ding{51}$^*$ & \ding{51} & \ding{55} & \ding{55} & \ding{55} & \ding{55} & \ding{55} \\
\midrule
\multicolumn{9}{l}{\textit{Conventional DL/ML approaches for power grid IDS}} \\
\midrule
Boyaci et al. \cite{boyaci2021chebyshev} (2021) & \ding{55} & \ding{55} & \ding{55} & \ding{51} & \ding{55} & \ding{55} & \ding{51} & \ding{55} \\
Niu et al. \cite{niu2019dynamic} (2019) & \ding{55} & \ding{55} & \ding{55} & \ding{51} & \ding{55} & \ding{55} & \ding{51} & \ding{55} \\
Lozano et al. \cite{lozano2026goose} (2026) & \ding{55} & \ding{51} & \ding{55} & \ding{55} & \ding{55} & \ding{55} & \ding{51} & \ding{55} \\
Herath et al. \cite{herath2025iec61850} (2025) & \ding{55} & \ding{51} & N/A & \ding{55} & \ding{55} & \ding{55} & \ding{51} & \ding{55} \\
\midrule
\multicolumn{9}{l}{\textit{Adversarial robustness of IDS}} \\
\midrule
Apruzzese et al. \cite{apruzzese2022modeling} (2022) & \ding{55} & \ding{51} & \ding{51} & \ding{55} & \ding{55} & \ding{55}$^\dagger$ & \ding{55} & \ding{55} \\
Cohen et al. \cite{cohen2019certified} (2019) & \ding{55} & \ding{51} & \ding{51} & \ding{55} & \ding{55} & \ding{55} & \ding{55} & \ding{55} \\
\midrule
\textbf{Ours (2026)} & \ding{51} & \ding{51} & \ding{51} & \ding{51} & \ding{51} & \ding{51} & \ding{51} & \ding{51} \\
\bottomrule
\multicolumn{9}{l}{\footnotesize $^*$General network IDS datasets (NSL-KDD, UNSW-NB15), not power-grid-specific.} \\
\multicolumn{9}{l}{\footnotesize $^\dagger$Evaluates evasion attacks (inference-time), not training-time poisoning/unlearning.} \\
\end{tabular}
\end{table*}

\section{Background}
\label{sec:preliminaries}
This section covers the fundamentals background of the proposed approach. We will cover three concepts, which are Leaky Integrate-and-Fire Neuron Model and Surrogate Gradient Learning, and machine unlearning.
\subsection{Leaky Integrate-and-Fire Neuron Model}

The Leaky Integrate-and-Fire (LIF) neuron is the fundamental computational unit in our framework. Unlike the Integrate-and-Fire neuron used in SNNPG \cite{hood2024snnpg}, the LIF model includes a membrane leak term that provides temporal memory with exponential decay. The membrane potential $U[t]$ evolves according to the discrete-time recurrence:
\begin{equation}
    U[t] = \beta \cdot U[t-1] + W \cdot I[t] - S[t-1] \cdot U_{\text{thr}}
    \label{eq:lif}
\end{equation}
where $\beta \in (0,1)$ is the membrane decay constant, $W$ is the synaptic weight matrix, $I[t]$ is the input current at timestep $t$, $S[t-1]$ is the spike output from the previous timestep, and $U_{\text{thr}}$ is the firing threshold. A spike is emitted when the membrane potential exceeds the threshold:
\begin{equation}
    S[t] = \Theta(U[t] - U_{\text{thr}})
    \label{eq:spike}
\end{equation}
where $\Theta(\cdot)$ denotes the Heaviside step function. The decay constant $\beta$ determines the neuron's temporal window. Values close to 1.0 retain information across many timesteps, while smaller values respond rapidly but forget prior evidence. We use $\beta = 0.85$ throughout, which balances temporal integration and responsiveness for the 15-timestep window in our framework.

\subsection{Surrogate Gradient Learning}

The Heaviside function in (\ref{eq:spike}) has zero gradient almost everywhere, which prevents standard backpropagation. We use surrogate gradient learning \cite{neftci2019surrogate}. This replaces the true gradient with a smooth approximation during the backward pass while retaining the discrete spike in the forward pass:
\begin{equation}
    \frac{\partial S}{\partial U} \approx \sigma'(k \cdot (U - U_{\text{thr}})) = \frac{k}{(1 + k|U - U_{\text{thr}}|)^2}
    \label{eq:surrogate}
\end{equation}
where $k = 25$ controls the sharpness of the fast sigmoid surrogate. This enables Backpropagation Through Time (BPTT) across all timesteps, directly optimising the cross-entropy loss through gradient descent. Unlike evolutionary approaches \cite{hood2024snnpg} that scale poorly with network size, or STDP-based methods \cite{mia2025neuromorphic} that optimise local synaptic objectives without global task alignment, surrogate gradient BPTT provides the scalability of modern deep learning while preserving spike-based computation for neuromorphic deployment.
\subsection{Machine Unlearning Attack Model}

We adopt the machine unlearning attack model proposed by Paphitis et al. \cite{paphitis2026unlearning}. Let $\mathcal{D}_{\text{train}} = \{(\mathbf{x}_i, y_i)\}_{i=1}^N$ be the training dataset with $C$ classes. The adversary targets a class $c^* \in \{1, \ldots, C\}$ and selects a fraction $\alpha \in [0, 1]$ of samples belonging to class $c^*$. For each selected sample, the label is replaced by a uniformly random label from $\{1, \ldots, C\} \setminus \{c^*\}$:
\begin{equation}
    \tilde{y}_i = \begin{cases}
    y_i & \text{if } y_i \neq c^* \text{ or } i \notin \mathcal{S}_\alpha \\
    \text{Uniform}(\{1,\ldots,C\} \setminus \{c^*\}) & \text{otherwise}
    \end{cases}
    \label{eq:attack}
\end{equation}
where $\mathcal{S}_\alpha$ is a random subset containing $\lfloor \alpha \cdot |\{i : y_i = c^*\}| \rfloor$ indices. The IDS model is then retrained on the corrupted dataset $\tilde{\mathcal{D}} = \{(\mathbf{x}_i, \tilde{y}_i)\}_{i=1}^N$. The attack succeeds if the retrained model's recall for class $c^*$ drops significantly while other metrics remain reasonable.

\section{Proposed Method}
\label{sec:proposed_method}
We developed the proposed architecture in way that has two piplines,  feature extraction branch and classification branch. 
We formulate the defence problem as follows. Let $\mathcal{M} = f_\theta \circ g_\phi$ denote an IDS model composed of a feature extractor $g_\phi: \mathbb{R}^d \rightarrow \mathbb{R}^m$ parameterised by $\phi$ and a classifier $f_\theta: \mathbb{R}^m \rightarrow \{1, \ldots, C\}$ parameterised by $\theta$. In standard end-to-end training, both $\phi$ and $\theta$ are updated jointly via:
\begin{equation}
    \phi^*, \theta^* = \arg\min_{\phi, \theta} \sum_{i=1}^N \mathcal{L}(f_\theta(g_\phi(\mathbf{x}_i)), y_i)
    \label{eq:joint_opt}
\end{equation}
When labels $y_i$ are corrupted to $\tilde{y}_i$, gradient updates propagate the corruption into \textit{both} the feature representation $g_\phi$ and the decision boundaries $f_\theta$. The feature space geometry itself becomes distorted, as samples of the targeted class are pulled toward other class clusters in the learned embedding space.

\begin{figure*}[!ht]
\centering
\includegraphics[width=\textwidth]{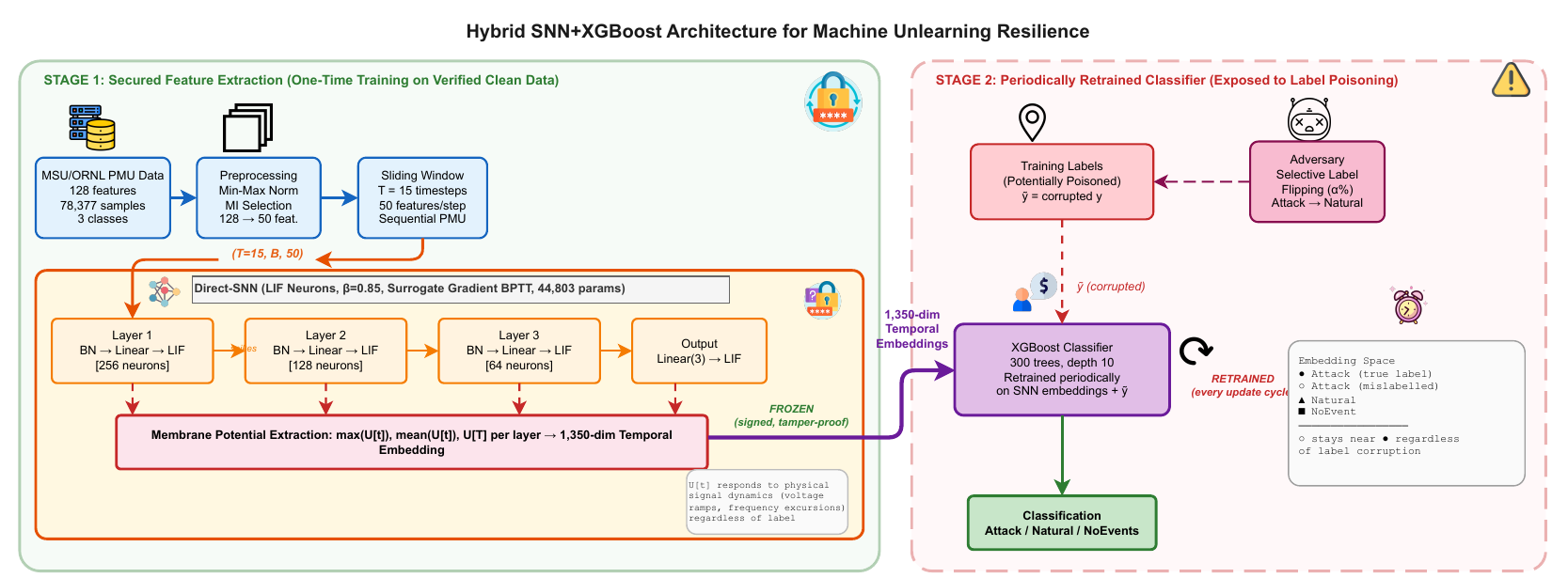}
\caption{Proposed hybrid SNN+XGBoost architecture for machine unlearning resilience. The SNN feature extractor (Stage~1) is trained once on verified clean data and deployed as a fixed neuromorphic frontend. Only the XGBoost classifier (Stage~2) is exposed to periodic retraining with potentially poisoned labels. The SNN temporal embeddings encode physical signal dynamics that are invariant to label corruption.}
\label{fig:framework}
\end{figure*}

We propose a different decomposition that separates the model into two components with distinct training regimes:
\begin{align}
    \text{Stage 1:} \quad & \phi^* = \arg\min_\phi \sum_{i=1}^{N_0} \mathcal{L}_{\text{SNN}}(g_\phi(\mathbf{x}_i), y_i^{\text{clean}}) \label{eq:stage1} \\
    \text{Stage 2:} \quad & \theta^* = \arg\min_\theta \sum_{i=1}^{N_t} \mathcal{L}_{\text{XGB}}(f_\theta(g_{\phi^*}(\mathbf{x}_i)), \tilde{y}_i) \label{eq:stage2}
\end{align}
where $\phi^*$ is computed once on verified clean data $\mathcal{D}_0 = \{(\mathbf{x}_i, y_i^{\text{clean}})\}_{i=1}^{N_0}$ and then frozen. Subsequent retraining updates only $\theta$ while the feature extractor remains fixed at $\phi^*$. The key insight is that label corruption in Stage 2 can only shift decision boundaries in the embedding space; it cannot alter the embedding space geometry itself. We illustrate the complete two-stage architecture in
in Fig.~\ref{fig:framework}.

The proposed approach consist of two stages as follows:

\textbf{Stage 1: Secured SNN Feature Extractor.} We train the Direct-SNN during system commissioning on a verified clean dataset. The SNN processes sliding windows of $T = 15$ consecutive PMU measurements and produces 1,350-dimensional temporal membrane potential embeddings. Once trained to convergence, we freeze all SNN parameters $\phi^* = \theta_{\text{SNN}}$ and cryptographically sign the parameter checkpoint using SHA-256 hashing. The signed hash is stored on a hardware security module at the substation, enabling tamper detection before each inference cycle. The frozen SNN then operates as a fixed neuromorphic frontend that transforms raw PMU time-series into structurally meaningful temporal representations. The geometry of these representations is determined by the physical dynamics of the power system, not by training labels.

\textbf{Stage 2: Periodically Retrained Classifier.} An XGBoost gradient-boosted decision tree classifier operates on the 1,350-dimensional SNN embeddings. This is the only component subject to periodic retraining (e.g., monthly or quarterly, as dictated by operational requirements) and therefore the only component exposed to potentially corrupted training labels. Because XGBoost learns axis-aligned splits on a fixed feature space, its sensitivity to label noise is bounded by the structure of that space. When the embedding space clusters samples by physical behaviour rather than label assignment, mislabelled samples appear as outliers that do not systematically shift tree boundaries.

\subsection{Data Preprocessing Pipeline}

We design the preprocessing pipeline to reduce dimensionality while preserving the temporal structure that is necessary for SNN membrane integration.

\textbf{Min-Max Normalisation.} We normalise each feature $x^{(j)}$ to the unit interval using training set statistics:
\begin{equation}
    \hat{x}_i^{(j)} = \frac{x_i^{(j)} - \min_{\text{train}}(x^{(j)})}{\max_{\text{train}}(x^{(j)}) - \min_{\text{train}}(x^{(j)}) + \epsilon}
    \label{eq:minmax}
\end{equation}
where $\epsilon = 10^{-8}$ prevents division by zero for constant features. We compute statistics on the training set and apply the same transformation to validation and test sets. Min-Max normalisation is preferred over standardisation because the resulting values in $[0, 1]$ directly map to biologically plausible input currents for LIF neurons without requiring additional scaling.

\textbf{Mutual Information Feature Selection.} For the MSU/ORNL dataset (128 raw features), we reduce dimensionality from 128 to $d = 50$ features using mutual information (MI) between each feature and the class label:
\begin{equation}
    \text{MI}(X^{(j)}; Y) = \sum_{x \in X^{(j)}} \sum_{y \in Y} p(x, y) \log \frac{p(x, y)}{p(x) \, p(y)}
    \label{eq:mi}
\end{equation}
We select the top-$d$ features ranked by MI score. MI is preferred over linear correlation because it captures nonlinear dependencies between PMU measurements and fault types. It is preferred over Principal Component Analysis (PCA) because MI preserves individual feature identities, which is important for physical interpretability, and because PCA maximises variance rather than class discrimination. In practice, MI selection retains all voltage magnitude, frequency, and ROCOF features while discarding redundant phasor angle combinations.

\textbf{Sliding Window Construction.} We construct temporal windows from consecutive PMU measurements to provide genuine time-varying input to the SNN:
\begin{equation}
    \mathbf{X}_{\text{win}}^{(i)} = [\hat{\mathbf{x}}_i, \hat{\mathbf{x}}_{i+1}, \ldots, \hat{\mathbf{x}}_{i+T-1}]^\top \in \mathbb{R}^{T \times d}
    \label{eq:window}
\end{equation}
with stride 1 (fully overlapping windows) and window size $T = 15$ timesteps. The label assigned to each window is the label of the final timestep: $y_{\text{win}}^{(i)} = y_{i+T-1}$. The choice of $T = 15$ is motivated by the LIF membrane time constant: with $\beta = 0.85$, the effective time constant is $\tau = -1 / \ln(\beta) \approx 6.1$ timesteps. A window of $T = 15$ provides approximately 2.4 membrane time constants, allowing the membrane potential to complete roughly two full integration cycles and capture both transient onset dynamics and sustained-state behaviour.

The preprocessing pipeline transforms raw data as follows: $128$ raw PMU features $\xrightarrow{\text{MI}}$ $50$ selected features $\xrightarrow{\text{window}}$ $\mathbf{X}_{\text{win}} \in \mathbb{R}^{15 \times 50}$ (750 values per sample, with temporal structure preserved across the first axis).
\subsection{Direct-SNN Architecture}

We propose the Direct-SNN architecture for power grid intrusion detection. The key idea is to bypass spike encoding entirely and treat normalised features as continuous input currents injected into LIF neurons at every timestep. This design is motivated by the observation that traditional spike encoding (rate, latency, delta modulation, etc.) introduces an information bottleneck for flat power system features, as we demonstrate in the ablation study (Section~\ref{sec:experiments}).
Given a sliding window $\mathbf{X}_{\text{win}} \in \mathbb{R}^{T \times d}$, each timestep provides a distinct input:
\begin{equation}
    I_l[t] = \text{BN}_l(W_l \cdot \mathbf{X}_{\text{win}}[t])
    \label{eq:direct_cur}
\end{equation}

The LIF neurons integrate these currents through their membrane dynamics:
\begin{equation}
    S_l[t], \; U_l[t] = \text{LIF}_l(I_l[t], \; U_l[t-1])
    \label{eq:direct_lif}
\end{equation}

The porposed SNN architecture consists of three hidden layers with [256, 128, 64] neurons, BatchNorm before each LIF layer, and dropout ($p = 0.2$). The total parameter count is 44,803,  which is 10 to 100 times smaller than typical CNN or LSTM-based IDS models. Sliding windows of size $T = 15$ over $d = 50$ features (selected from 128 via mutual information) provide genuine temporal variation. Unlike spike encoding, which binarises feature values and destroys magnitude information, direct injection preserves the full continuous values. Subtle differences in feature magnitudes (e.g., $x_1 = 0.72$ versus $x_1 = 0.74$) produce different membrane potential trajectories across layers and timesteps. The SNN can therefore discriminate between attack types based on fine-grained measurement differences that spike encoding would destroy.

\subsection{Temporal Membrane Embedding Extraction}

For each hidden layer $l$ with $N_l$ neurons, we record the membrane trajectory $\{U_l[t]\}_{t=1}^T$ and extract:
\begin{align}
    \mathbf{f}_l^{\max} &= \max_{t \in [1,T]} U_l[t] \in \mathbb{R}^{N_l} \\
    \mathbf{f}_l^{\text{mean}} &= \frac{1}{T}\sum_{t=1}^T U_l[t] \in \mathbb{R}^{N_l} \\
    \mathbf{f}_l^{\text{last}} &= U_l[T] \in \mathbb{R}^{N_l}
\end{align}
The full embedding concatenates all layer features:
\begin{equation}
    \mathbf{e} = [\mathbf{f}_1^{\max}; \mathbf{f}_1^{\text{mean}}; \mathbf{f}_1^{\text{last}}; \ldots; \mathbf{f}_L^{\max}; \mathbf{f}_L^{\text{mean}}; \mathbf{f}_L^{\text{last}}; \mathbf{f}_{\text{out}}^{\max}; \mathbf{f}_{\text{out}}^{\text{mean}}]
    \label{eq:embedding}
\end{equation}
yielding a 1,350-dimensional embedding for the [256, 128, 64] architecture with $C = 3$ output classes.

\subsection{XGBoost Classifier on Temporal Embeddings}
\label{sec:xgboost}
We select XGBoost (eXtreme Gradient Boosting) as the Stage 2 classifier for four reasons. First, gradient-boosted trees handle high-dimensional structured features (1,350 dimensions) effectively without requiring explicit feature engineering. Second, XGBoost provides native feature importance rankings, enabling interpretability of which embedding dimensions drive classification decisions. Third, XGBoost retraining is computationally lightweight (4.7 seconds on CPU for 78,377 samples), making frequent model updates practical for operational deployments. Fourth, tree-based classifiers are inherently regularised through maximum depth constraints and subsampling, providing baseline noise tolerance.

We configure XGBoost with 300 boosting rounds, maximum tree depth 10, learning rate 0.1, column subsampling ratio 0.8, row subsampling ratio 0.8, and L2 regularisation $\lambda = 1.0$. The multi-class objective uses softmax with class weights inversely proportional to class frequency. These hyperparameters are selected via grid search on the validation set under clean (0\% attack) conditions and held fixed across all attack levels.

XGBoost receives the 1,350-dimensional embedding vector $\mathbf{e}$ as input features and the (potentially corrupted) label $\tilde{y}$ as the target. Each tree split operates on a single embedding dimension, partitioning the structured temporal feature space. Because the embedding dimensions encode physically meaningful quantities (peak membrane response of specific neurons to specific temporal patterns), each split has a physical interpretation.

\subsection{Why Temporal Embeddings Resist Label Poisoning}
\label{sec:mechanism}

The robustness mechanism operates through three complementary effects:

\textbf{(1) Physics-grounded clustering.} The SNN membrane dynamics respond to the temporal characteristics of the input signal. A cyber attack on the power grid produces distinctive voltage and current transients (rapid voltage drops, frequency excursions, anomalous phase angles) that differ physically from natural disturbances (load changes, switching events) and normal operation. These physical differences produce distinct membrane potential trajectories \textit{regardless of the assigned label}. In embedding space, attack samples cluster together because they share temporal dynamics, not because they share a label.

\textbf{(2) Label-independent feature extraction.} Because we train the SNN once on clean data and then freeze it, label corruption during periodic retraining \textit{cannot affect the embeddings}. The SNN continues to produce the same temporal representations for the same input signals. Only the XGBoost decision boundaries shift. This is fundamentally different from end-to-end trained models where label corruption corrupts both the representation and the classifier.

\textbf{(3) Structured embedding space.} XGBoost operating on the structured 1,350-dimensional embedding space is more robust to label noise than XGBoost on raw 750-dimensional flattened features. The SNN embeddings contain temporal derivative-like features (captured via max membrane), sustained-activity features (mean membrane), and integration features (last membrane state). These structured representations make the tree splits more meaningful and less susceptible to individual mislabelled points shifting boundaries. Algorithm~\ref{alg:deployment} summarises the complete forward pass and training loop..

\begin{algorithm}[t]
\caption{Secure Hybrid SNN+XGBoost Deployment}
\label{alg:deployment}
\begin{algorithmic}[1]
\REQUIRE Verified clean dataset $\mathcal{D}_{\text{clean}}$
\STATE \textbf{// Stage 1: One-time secured training}
\STATE Train Direct-SNN on $\mathcal{D}_{\text{clean}}$
\STATE Extract embeddings $\mathbf{E} = \text{SNN}(\mathcal{D}_{\text{clean}})$
\STATE Freeze $\theta_{\text{SNN}}$; deploy to neuromorphic hardware
\STATE Sign $\theta_{\text{SNN}}$ for tamper detection
\STATE
\STATE \textbf{// Stage 2: Periodic retraining (exposed to poisoning)}
\FORALL{retraining epochs}
    \STATE Collect new data $\mathcal{D}_{\text{new}}$ (labels may be corrupted)
    \STATE $\mathbf{E}_{\text{new}} \leftarrow \text{SNN}(\mathcal{D}_{\text{new}}; \theta_{\text{SNN}})$ \COMMENT{Fixed SNN}
    \STATE Retrain XGBoost on $(\mathbf{E}_{\text{new}}, \tilde{y})$
    \STATE Deploy updated classifier
\ENDFOR
\end{algorithmic}
\end{algorithm}



\section{Experimental setup}
\label{sec:experiments_setup}
In this section, We first introduce the datasets used in our experiments. This is followed by baseline and metrics used in our experiments. 
\subsection{Datasets}
We evaluate the proposed framework on two real public power grid cyberattack datasets, as summarized in Table~\ref{tab:dataset_summary}. Both datasets contain measurements from actual power system equipment. This distinguishes our evaluation from all prior SNN power grid work that relies on synthetic simulators.

\begin{table}[!t]
\centering
\footnotesize
\caption{Dataset Summary Statistics.}
\label{tab:dataset_summary}
\renewcommand{\arraystretch}{1.1}
\begin{tabular}{lcc}
\toprule
\textbf{Property} & \textbf{Synchrophasor} & \textbf{MSU/ORNL} \\
\midrule
Total samples & 218,459 & 78,377 \\
Raw features & 9 & 128 \\
Selected features & 9 (all) & 50 (MI) \\
Classes & 3 & 3 \\
Sampling rate & 30 Hz & 30 Hz \\
Attack types & FDI, TSA & FDI, TSA, Replay \\
Source & IEEE testbed & MSU/ORNL testbed \\
\midrule
\multicolumn{3}{l}{\textit{Class distribution}} \\
\quad Attack & 37.2\% & 71.0\% (55,663) \\
\quad Natural & 30.1\% & 23.4\% (18,309) \\
\quad NoEvents & 32.7\% & 5.6\% (4,405) \\
\bottomrule
\end{tabular}
\end{table}

\textbf{Synchrophasor Cyberattack Dataset}: The Synchrophasor Cyberattack Dataset~\cite{synchrophasor_dataset} contains PMU measurements from a real electrical distribution network under normal operation with two attacks scenarios. The dataset has 77,072 normal samples of the PMU measurements, 64,280 FDI attacks samples, and 77,107 TSA attack samples. The total number of samples is 218,459 across 9 features. Table~\ref{tab:synchro_features} details the feature set, and Table~\ref{tab:synchro_classes} describes the attack classes.
\begin{table}[t]
\centering
\footnotesize
\caption{Feature Description of the Synchrophasor PMU Dataset (9 Features)}
\label{tab:synchro_features}
\footnotesize
\begin{tabular}{lll}
\toprule
\textbf{Feature} & \textbf{Category} & \textbf{Description} \\
\midrule
$V_a$, $V_b$, $V_c$ & Voltage & Three-phase voltage magnitudes \\
$I_a$, $I_b$, $I_c$ & Current & Three-phase current magnitudes \\
$f$ & Frequency & System frequency (Hz) \\
ROCOF & Frequency & Rate-of-change-of-frequency \\
$\phi$ & Phase & Voltage-current phase angle \\
\bottomrule
\end{tabular}
\end{table}
The Synchrophasor dataset is well suited for validating classification accuracy because its 9 features are compact, non-redundant, and directly correspond to the fundamental electrical quantities that distinguish attack signatures from normal operation. Each feature carries independent discriminative information.

\begin{table}[t]
\centering
\caption{Attack Classes in the Synchrophasor Dataset}
\label{tab:synchro_classes}
\footnotesize
\begin{tabular}{lccp{4.2cm}}
\toprule
\textbf{Class} & \textbf{Samples} & \textbf{\%} & \textbf{Description} \\
\midrule
Normal & 77,072 & 35.3 & Legitimate PMU measurements under nominal grid operation \\
FDI & 64,280 & 29.4 & False Data Injection: corrupted measurements to mislead state estimation \\
TSA & 77,107 & 35.3 & Time Synchronisation Attack: GPS spoofing causing phasor misalignment \\
\bottomrule
\end{tabular}
\end{table}

\textbf{MSU/ORNL Power System Attack Dataset} \cite{msu_ornl}. This dataset contains 78,377 samples with 128 PMU features collected from a realistic power system testbed at Mississippi State University and Oak Ridge National Laboratory. The testbed instruments 4 protective relay locations, each providing voltage/current phasors across three phases, plus system-wide frequency and control signals. Table~\ref{tab:msu_features} describes the feature groups and their roles in attack detection.

The MSU/ORNL dataset is particularly suitable for machine unlearning evaluation for three reasons. First, the severe class imbalance (71\% Attack, 23\% Natural, 6\% NoEvents) mirrors real-world deployments where attacks may dominate training data during active campaigns. Second, the 128-feature high dimensionality introduces redundancy (many phasor pairs are correlated), making raw-feature classifiers more susceptible to label noise than those operating on a compressed representation. Third, distinguishing attacks from natural disturbances requires temporal pattern recognition. Both produce voltage and current transients, but attacks exhibit non-physical combinations (e.g., voltage magnitude change without corresponding current change, or frequency deviation without ROCOF correlation) that only become apparent when measurements are analysed as time-series rather than independent snapshots.

\textbf{Feature Selection.} Mutual information selection on the MSU/ORNL dataset retains the following feature groups in the top 50: all frequency and ROCOF features (high MI due to direct correlation with event type), voltage magnitudes from relays 1 and 3 (electrically close to the attack injection points), relay trip/close status (binary indicators of protection operation), and selected sequence components. Redundant phasor angle pairs and highly correlated inter-phase measurements are discarded, reducing noise without losing discriminative power.

\begin{table}[!t]
\centering
\footnotesize
\caption{MSU/ORNL Dataset Feature Groups (128 features).}
\label{tab:msu_features}
\renewcommand{\arraystretch}{1.05}
\begin{tabular}{p{2.8cm}p{3.9cm}}
\toprule
\textbf{Feature Group} & \textbf{Physical Significance} \\
\midrule
Voltage magnitudes ($|V_{a,b,c}|$) (12)
& Voltage profiles across relays; spatial correlation indicates localised or system-wide events \\

Current magnitudes ($|I_{a,b,c}|$) (12)
& Current-flow patterns used by directional protection logic \\

Voltage angles ($\angle V_{a,b,c}$) (12)
& Inter-relay angular differences support transient-stability monitoring \\

Current angles ($\angle I_{a,b,c}$) (12)
& Power-flow direction; abnormal reversal may indicate attack or islanding \\

Sequence components ($V_+,V_-,V_0$) (12)
& Positive-, negative-, and zero-sequence components quantify voltage unbalance \\

Frequency and ROCOF (8)
& Local frequency dynamics; inter-relay differences indicate loss of coherency \\

Relay trip/close status (16)
& Binary protection states; unexpected operations may indicate control-plane attacks \\

Active/reactive power (16)
& Real and reactive power flow; abrupt changes indicate generation or load events \\

Control panel logs (38)
& Breaker, tap-changer, and capacitor-bank states; anomalous switching may indicate attacks \\
\bottomrule
\end{tabular}
\end{table}

\subsection{Baselines and metrics}
We compare the proposed method framework against seven baseline methods. These methods are spanning from deep learning, such as MLP~\cite{takiddin2022mlp_grid}, Long Short-Term Memory (LSTM)\cite{wang2020lstm_grid}, 1D Convolutional Neural Network (1D-CNN)\cite{zhang2021cnn_grid}, and Graph Neural Network (GNN)\cite{boyaci2021chebyshev}, Tree-based ensemble, such as Random Forest (RF)\cite{ahmed2022rf_grid},XGBoost\cite{li2022xgboost_grid} and ruble based approaches, such as Rule-Based Detector\cite{herath2025iec61850}. The parameter setting of these methods is shown in Table \ref{tab:baseline_params}.

\begin{table}[!t]
\centering
\footnotesize
\caption{Hyperparameter Settings for All Methods. All deep learning models use Adam optimiser with class-weighted cross-entropy loss and batch size 256.}
\label{tab:baseline_params}
\renewcommand{\arraystretch}{1.2}
\begin{tabular}{lp{5.2cm}}
\toprule
\textbf{Method} & \textbf{Settings} \\
\midrule
MLP \cite{takiddin2022mlp_grid} & Layers: [256, 128, 64], LR: $10^{-3}$, dropout: 0.2, BatchNorm, epochs: 50 \\
\addlinespace
LSTM \cite{wang2020lstm_grid} & 2 layers, 128 hidden units, LR: $10^{-3}$, epochs: 50 \\
\addlinespace
1D-CNN \cite{zhang2021cnn_grid} & Filters: [64, 128, 256], kernel: 3, GAP, LR: $10^{-3}$, epochs: 50 \\
\addlinespace
GNN \cite{boyaci2021chebyshev} & 2 ChebConv layers, 64 units, $K=2$, LR: $10^{-3}$, epochs: 50 \\
\addlinespace
RF \cite{ahmed2022rf_grid} & 300 trees, max depth: 20, class weight: balanced \\
\addlinespace
XGBoost \cite{li2022xgboost_grid} & 300 rounds, depth: 10, LR: 0.1, colsample: 0.8, subsample: 0.8, $\lambda$: 1.0 \\
\addlinespace
Rule-Based \cite{herath2025iec61850} & $|V|$: $\pm$0.1 pu, $f$: $\pm$0.5 Hz, ROCOF: $\pm$1.0 Hz/s \\
\midrule
\textbf{Direct-SNN} & Layers: [256, 128, 64], LIF ($\beta$: 0.85, $U_\text{thr}$: 1.0), $T$: 15, LR: $5\times10^{-4}$, dropout: 0.2, epochs: 50 \\
\addlinespace
\textbf{Hybrid (Ours)} & SNN (frozen) + XGBoost (300 rounds, depth: 10, LR: 0.1, $\lambda$: 1.0) \\
\bottomrule
\end{tabular}
\end{table}

We use four metrics to report models performance: overall accuracy, F1-macro (equally weights all classes regardless of support), Matthews Correlation Coefficient (MCC, balanced under class imbalance), and per-class precision and recall. We also report parameter count as a proxy for neuromorphic deployability.
Overall accuracy measures the fraction of correctly classified samples:
\begin{equation}
    \text{Accuracy} = \frac{1}{N}\sum_{i=1}^{N} \mathbb{1}[\hat{y}_i = y_i]
    \label{eq:accuracy}
\end{equation}
where $N$ is the total number of test samples and $\mathbb{1}[\cdot]$ is the indicator function.

For each class $c$, precision and recall are defined as:
\begin{equation}
    \text{Precision}_c = \frac{\text{TP}_c}{\text{TP}_c + \text{FP}_c}, \quad
    \text{Recall}_c = \frac{\text{TP}_c}{\text{TP}_c + \text{FN}_c}
    \label{eq:prec_rec}
\end{equation}
where $\text{TP}_c$, $\text{FP}_c$, and $\text{FN}_c$ denote true positives, false positives, and false negatives for class $c$, respectively. The per-class F1 score is the harmonic mean of precision and recall. We report F1-macro, which weights all classes equally regardless of support:
\begin{equation}
    \text{F1-macro} = \frac{1}{C}\sum_{c=1}^{C} \frac{2 \cdot \text{Precision}_c \cdot \text{Recall}_c}{\text{Precision}_c + \text{Recall}_c}
    \label{eq:f1macro}
\end{equation}

The Matthews Correlation Coefficient (MCC) provides a balanced
measure of classification quality and is particularly useful under
class imbalance. For a multi-class confusion matrix $\mathbf{C}$,
MCC is defined as
\begin{equation}
\mathrm{MCC} =
\frac{
cs-\sum_{k=1}^{K}p_k t_k
}{
\sqrt{
\left(s^2-\sum_{k=1}^{K}p_k^2\right)
\left(s^2-\sum_{k=1}^{K}t_k^2\right)
}
},
\label{eq:mcc}
\end{equation}
where $s$ is the total number of samples, $c$ is the number of
correct predictions, $p_k$ is the number of samples with true class
$k$, and $t_k$ is the number of samples predicted as class $k$.
MCC has a maximum value of $+1$ for perfect classification and
provides a correlation-based measure that remains informative under
class imbalance.
\subsection{Implementation details}

\textbf{Preprocessing.} For both datasets, we apply Min-Max normalisation. For MSU/ORNL, mutual information feature selection reduces dimensionality from 128 to 50 features. We construct sliding windows of size $T = 15$ from consecutive samples. Data is split 70/15/15 (train/val/test) for baseline evaluation and 70/30 (train/test) for unlearning experiments, with fixed random seeds.

\textbf{SNN Architecture.} The Direct-SNN uses [256, 128, 64] hidden layers with LIF neurons ($\beta = 0.85$, $U_{\text{thr}} = 1.0$), BatchNorm before each layer, dropout ($p = 0.2$), and 15 simulation timesteps. Total parameters: 44,803. Training uses surrogate gradient BPTT via snnTorch with learning rate 0.0005, batch size 256, and balanced class weights.


All models are implemented in PyTorch 2.5 with snnTorch 0.9 on NVIDIA CUDA. We use a fixed random seed (42) with a 70/15/15 train/validation/test split applied identically across all experiments. Features are normalised to $[0,1]$ using training set statistics. All spike-encoded models use $T = 25$ timesteps. The Direct SNN uses $T = 15$ timesteps (determined via hyperparameter search).

\section{Results and Discussion}
\label{sec:results}
\subsection{Classification Results}
We first establish that the hybrid architecture achieves strong classification accuracy on clean data before evaluating robustness. Table~\ref{tab:hybrid_results} presents the hybrid SNN+XGBoost results across both datasets compared to standalone baselines.

\begin{table*}[!t]
\centering
\caption{Hybrid SNN+XGBoost Classification Results on Clean Data Across Two Real Public Power Grid Datasets. The hybrid architecture outperforms both standalone SNN and standalone tree models on PMU-based datasets.}
\label{tab:hybrid_results}
\renewcommand{\arraystretch}{1.15}
\begin{tabular}{llcccl}
\toprule
\textbf{Dataset} & \textbf{Method} & \textbf{Accuracy} & \textbf{F1-Macro} & \textbf{MCC} & \textbf{Embedding Dim.} \\
\midrule
\multirow{5}{*}{\textbf{Synchrophasor (9 feat.)}} & Standalone SNN & 65.0\% & 0.548 & 0.606 & N/A \\
& Raw XGBoost & 95.7\% & 0.955 & 0.937 & N/A \\
& Raw Random Forest & 95.5\% & 0.953 & 0.934 & N/A \\
& \textbf{Hybrid SNN+XGBoost} & \textbf{99.9\%} & \textbf{0.999} & \textbf{0.998} & 1,350 \\
& Hybrid SNN+RF & 99.7\% & 0.997 & 0.995 & 1,350 \\
\midrule
\multirow{5}{*}{\textbf{MSU/ORNL (128 feat.)}} & Standalone SNN & 66.6\% & 0.655 & 0.445 & N/A \\
& Raw XGBoost & 90.1\% & 0.883 & 0.768 & N/A \\
& Raw Random Forest & 87.9\% & 0.842 & 0.715 & N/A \\
& \textbf{Hybrid SNN+XGBoost} & \textbf{95.0\%} & \textbf{0.943} & \textbf{0.885} & 1,350 \\
& Hybrid SNN+RF & 87.9\% & 0.850 & 0.715 & 1,350 \\
\bottomrule
\end{tabular}
\end{table*}

The proposed hybrid achieves 99.9\% accuracy on the Synchrophasor dataset (+4.2\% over raw XGBoost) and 95.0\% on MSU/ORNL (+4.9\% over raw XGBoost). These results demonstrate that SNN temporal embeddings provide complementary information to raw features.

Table~\ref{tab:baselines_synchro} compares all baseline methods on the Synchrophasor dataset, and Table~\ref{tab:baselines_msu} on the MSU/ORNL dataset.

\begin{table}[!t]
\centering
\caption{Baseline Comparison on Synchrophasor PMU Dataset (3-Class, 218K Samples, 70/15/15 Split)}
\label{tab:baselines_synchro}
\begin{tabular}{lcccc}
\toprule
\textbf{Method} & \textbf{Acc.} & \textbf{F1-M} & \textbf{MCC} & \textbf{Params} \\
\midrule
MLP \cite{takiddin2022mlp_grid} & 70.53\% & 0.57 & 0.00 & 44,803 \\
LSTM \cite{wang2020lstm_grid} & 70.53\% & 0.57 & 0.00 & 203,651 \\
1D-CNN \cite{zhang2021cnn_grid} & 70.53\% & 0.57 & 0.00 & 157,059 \\
GNN \cite{boyaci2021chebyshev} & 70.53\% & 0.57 & 0.00 & 82,563 \\
Random Forest \cite{ahmed2022rf_grid} & 95.41\% & 0.95 & 0.92 & N/A \\
XGBoost \cite{li2022xgboost_grid} & 95.74\% & 0.96 & 0.93 & N/A \\
Rule-Based \cite{herath2025iec61850} & 35.18\% & 0.17 & 0.04 & N/A \\
\midrule
Direct-SNN (Ours) & 93.5\% & 0.93 & 0.89 & 44,803 \\
\textbf{Hybrid SNN+XGB (Ours)} & \textbf{99.9\%} & \textbf{0.999} & \textbf{0.998} & 44,803+XGB \\
\bottomrule
\end{tabular}
\end{table}

\begin{table}[!t]
\centering
\caption{Baseline Comparison on MSU/ORNL Power System Attack Dataset (128 Features $\rightarrow$ 50 via MI Selection, 78K Samples, 3 Classes)}
\label{tab:baselines_msu}
\begin{tabular}{lccc}
\toprule
\textbf{Method} & \textbf{Accuracy} & \textbf{F1-Macro} & \textbf{MCC} \\
\midrule
MLP \cite{takiddin2022mlp_grid} & 58.6\% & 0.55 & 0.31 \\
LSTM \cite{wang2020lstm_grid} & 51.3\% & 0.42 & 0.15 \\
1D-CNN \cite{zhang2021cnn_grid} & 58.9\% & 0.52 & 0.28 \\
GNN \cite{boyaci2021chebyshev} & 55.4\% & 0.47 & 0.21 \\
Direct-SNN (standalone) & 66.6\% & 0.66 & 0.45 \\
\midrule
Random Forest \cite{ahmed2022rf_grid} & 87.9\% & 0.84 & 0.72 \\
XGBoost \cite{li2022xgboost_grid} & 90.1\% & 0.88 & 0.77 \\
Rule-Based \cite{herath2025iec61850} & 66.4\% & 0.27 & 0.04 \\
\midrule
\textbf{Hybrid SNN+XGBoost (Ours)} & \textbf{95.0\%} & \textbf{0.94} & \textbf{0.89} \\
\bottomrule
\end{tabular}
\end{table}

All four deep learning baselines (MLP, LSTM, 1D-CNN, GNN) converge to majority-class accuracy on both datasets due to the rate encoding bottleneck we identified in Section \ref{sec:encoding_ablation}. Tree-based methods perform well but lack temporal modelling. The proposed hybrid combines the strengths of both approaches: SNN temporal embeddings for feature extraction with tree-based classification for decision making.

\subsection{Robustness to label poisoning and machine unlearning}

In this section, we now evaluate its robustness under the machine unlearning attack on the MSU/ORNL dataset.

\textbf{Attack Setup.} Following Paphitis et al. \cite{paphitis2026unlearning}, we target Class~0 (Attack) as the unlearning target. At each level $\alpha \in \{0, 10, 20, 30, 40, 50, 60, 70, 80, 90\}\%$, we randomly relabel $\alpha\%$ of Attack-class training samples to Natural or NoEvents. Test labels remain unmodified. We retrain all models from scratch at each attack level. The SNN feature extractor remains frozen; it is never exposed to poisoned labels.

In this section, we evalaute the performanc of the following methods:
\begin{itemize}
    \item Hybrid SNN+XGBoost: XGBoost (300 trees, depth 10) on 1,350-dim SNN embeddings.
    \item Raw XGBoost: XGBoost (300 trees, depth 10) on 750-dim flattened window features.
    \item Raw Random Forest: RF (300 trees, depth 20) on 750-dim flattened features.
\end{itemize}

We report the following metrics: F1-Macro across all classes; Target-class F1 (F1 of the attacked class);  Target-class Recall (detection rate); and  Degradation from baseline (relative F1-Macro loss compared to 0\% attack). We define the \textit{collapse threshold} as the attack percentage where target-class F1 drops below 0.5.


Table~\ref{tab:main_results} presents the full robustness comparison across all attack levels. We also summarise the key robustness metrics in Table~\ref{tab:robustness_summary}. 

\begin{table*}[!t]
\centering
\caption{Machine Unlearning Robustness Comparison on MSU/ORNL Dataset. Target class: Attack (Class 0). F1-M = F1-Macro, Tgt-F1 = Target-class F1, Tgt-Rec = Target-class Recall. Bold indicates best performance at each attack level.}
\label{tab:main_results}
\renewcommand{\arraystretch}{1.15}
\begin{tabular}{c|ccc|ccc|ccc}
\toprule
\textbf{Attack} & \multicolumn{3}{c|}{\textbf{Hybrid SNN+XGBoost}} & \multicolumn{3}{c|}{\textbf{Raw XGBoost}} & \multicolumn{3}{c}{\textbf{Raw Random Forest}} \\
\textbf{Level} & F1-M & Tgt-F1 & Tgt-Rec & F1-M & Tgt-F1 & Tgt-Rec & F1-M & Tgt-F1 & Tgt-Rec \\
\midrule
0\% & \textbf{0.928} & \textbf{0.959} & 0.988 & 0.899 & 0.939 & 0.982 & 0.755 & 0.890 & \textbf{0.994} \\
10\% & \textbf{0.919} & \textbf{0.954} & \textbf{0.972} & 0.900 & 0.938 & 0.963 & 0.796 & 0.900 & 0.987 \\
20\% & \textbf{0.900} & \textbf{0.942} & \textbf{0.938} & 0.893 & 0.932 & 0.932 & 0.838 & 0.911 & 0.969 \\
30\% & \textbf{0.861} & \textbf{0.914} & \textbf{0.875} & 0.857 & 0.902 & 0.860 & 0.856 & 0.907 & 0.915 \\
40\% & 0.798 & \textbf{0.861} & \textbf{0.773} & 0.794 & 0.846 & 0.753 & \textbf{0.822} & 0.862 & 0.793 \\
50\% & 0.667 & \textbf{0.728} & \textbf{0.578} & 0.669 & 0.712 & 0.558 & \textbf{0.691} & 0.683 & 0.522 \\
60\% & \textbf{0.511} & \textbf{0.503} & \textbf{0.337} & 0.503 & 0.469 & 0.307 & 0.463 & 0.288 & 0.168 \\
70\% & \textbf{0.367} & \textbf{0.225} & \textbf{0.127} & 0.363 & 0.197 & 0.109 & 0.354 & 0.045 & 0.023 \\
80\% & 0.296 & \textbf{0.062} & \textbf{0.032} & 0.293 & 0.046 & 0.024 & \textbf{0.331} & 0.004 & 0.002 \\
90\% & 0.276 & \textbf{0.010} & \textbf{0.005} & 0.277 & 0.007 & 0.003 & \textbf{0.329} & 0.001 & 0.001 \\
\bottomrule
\end{tabular}
\end{table*}


\begin{table}[!t]
\centering
\caption{Robustness Summary Metrics. AUDC = Average Degradation across all attack levels (lower is better). Collapse = attack \% where target-class F1 drops below 0.5.}
\label{tab:robustness_summary}
\begin{tabular}{lccc}
\toprule
\textbf{Metric} & \textbf{Hybrid} & \textbf{Raw XGB} & \textbf{Raw RF} \\
\midrule
Baseline F1-Macro & \textbf{0.928} & 0.899 & 0.755 \\
Baseline Target-F1 & \textbf{0.959} & 0.939 & 0.890 \\
Degrad. @ 10\% & \textbf{0.9\%} & $-$0.1\% & $-$5.5\% \\
Degrad. @ 30\% & 7.3\% & \textbf{4.7\%} & $-$13.3\% \\
Degrad. @ 50\% & 28.1\% & \textbf{25.6\%} & \textbf{8.5\%} \\
AUDC & 29.7\% & 28.3\% & \textbf{17.4\%} \\
Collapse threshold & \textbf{70\%} & 60\% & 60\% \\
Target recall @ 90\% & \textbf{0.005} & 0.003 & 0.001 \\
\bottomrule
\end{tabular}
\end{table}
We highlight four key findings from our robustness evaluation:
\textbf{(1) Superior robustness at operationally critical levels.} At 10\% attack (the threshold Paphitis et al. identify as achievable with minimal adversary effort), the hybrid maintains target-class F1 of 0.954, representing only 0.5\% degradation from its baseline of 0.959. Raw XGBoost remains at 0.938, and RF achieves 0.900. The hybrid's \textit{absolute} target-class performance dominates at all levels from 0\% through 70\%.

\textbf{(2) Delayed collapse.} The hybrid's target-class F1 remains above 0.5 until 70\% poisoning, compared to 60\% for both raw models. This 10-percentage-point delay in functional collapse represents a significant additional resilience margin for operational deployments.

\textbf{(3) Consistent target-class advantage.} At every attack level from 0\% through 70\%, the hybrid achieves the highest target-class F1 and recall. This advantage is most pronounced at moderate attack levels (20\% to 50\%) where early detection of the unlearning attack is critical.

\textbf{(4) Baseline accuracy advantage.} Even without any attack, the hybrid achieves 0.928 F1-macro compared to 0.899 for raw XGBoost, a 3.2\% improvement attributable to the richer temporal embedding representation. This means the hybrid starts from a higher baseline, providing additional buffer against degradation.

\begin{figure*}[!t]
    \centering
    \subfloat[F1-Macro degradation across attack levels.]
    {
        \includegraphics[width=0.48\textwidth]{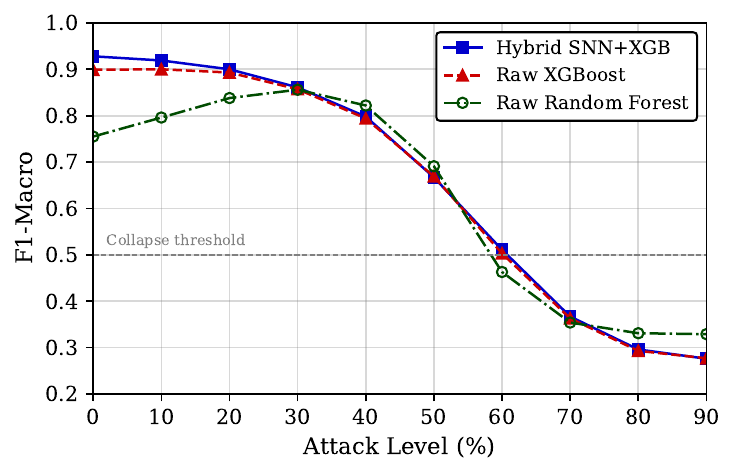}
        \label{fig:f1macro_line}
    }
    \hfill
    \subfloat[Target-class F1 showing collapse delay.]
    {
        \includegraphics[width=0.48\textwidth]{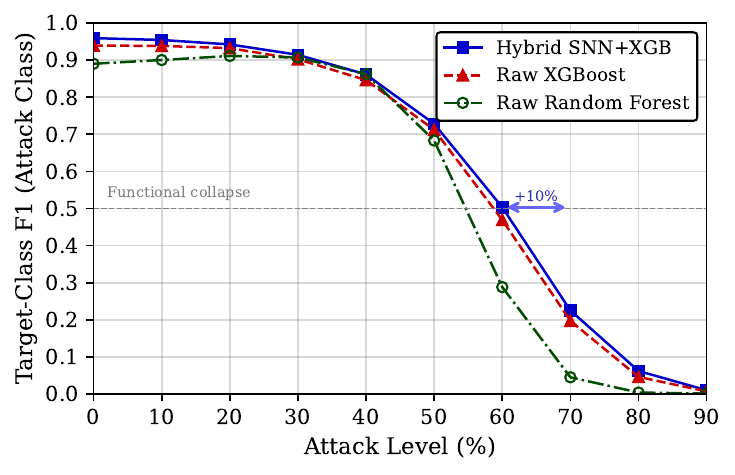}
        \label{fig:target_f1_line}
    }
    \caption{Degradation curves under increasing machine unlearning attack on the MSU/ORNL dataset. (a) Overall F1-Macro shows the hybrid maintains superior performance through 60\% poisoning. (b) Target-class F1 reveals the hybrid delays functional collapse (F1~$<$~0.5) by 10 percentage points compared to raw models. The shaded annotation highlights the collapse delay window.}
    \label{fig:degradation_curves}
\end{figure*}

\begin{figure}[!t]
\centering
\includegraphics[width=0.48\textwidth]{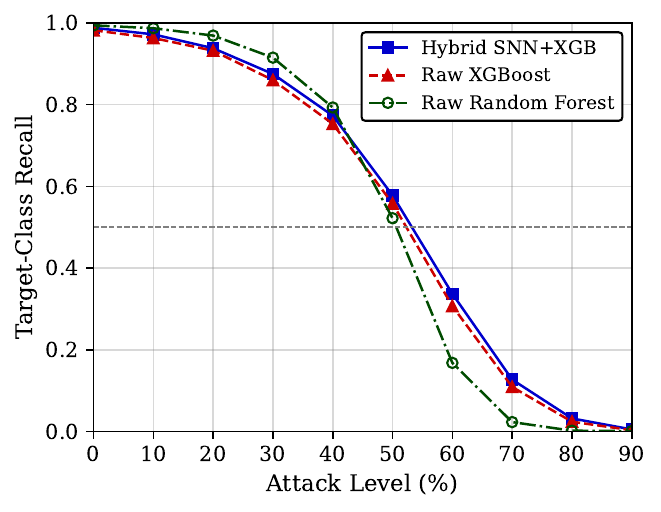}
\caption{Target-class recall (detection rate) under increasing attack. The hybrid consistently detects a higher fraction of true attack events, with the advantage most pronounced at 50--70\% poisoning where RF recall collapses rapidly.}
\label{fig:target_recall}
\end{figure}

\begin{figure}[!t]
\centering\includegraphics[width=0.48\textwidth]{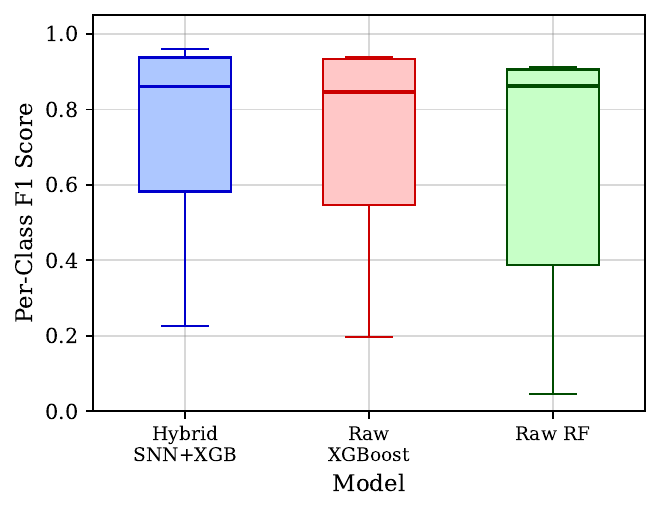}
\caption{Distribution of target-class F1 scores across attack levels (0--70\%). The hybrid exhibits the highest median and tightest interquartile range, demonstrating more consistent robustness. Raw Random Forest shows the largest spread, reflecting its catastrophic collapse at moderate attack levels.}
\label{fig:boxplot_targetf1}
\end{figure}

Our results provide empirical evidence for a broader principle: \textit{physics-grounded feature extractors provide inherent robustness against label-level attacks}. The SNN membrane potential dynamics are governed by the physical characteristics of the input signal, specifically the temporal patterns of voltage ramps, current transients, and frequency excursions. These dynamics are determined by the \textit{measurement} (i.e., the physics of the grid), not the \textit{label} (i.e., the human annotation). When an adversary corrupts labels, they cannot change the physical measurements and therefore cannot change the SNN embeddings.

This is analogous to how a spectrogram computed from an audio signal clusters phonetically similar sounds together regardless of transcription errors. The spectrogram encodes acoustic physics; the SNN embedding encodes electrical physics. Both are robust to annotation noise because they operate below the label abstraction.


The Raw Random Forest shows anomalous behaviour at low attack levels: its F1-macro \textit{improves} from 0.755 at 0\% to 0.856 at 30\%. This occurs because the RF baseline performs poorly on the high-dimensional input (750 features from flattened windows), and label noise acts as an implicit regulariser that prevents overfitting to the majority class. However, this improvement is illusory. The RF target-class recall drops steadily, and the model collapses catastrophically at 60\% attack. The robustness of the proposed hybrid, by contrast, stems from structural properties of the embedding space rather than accidental regularisation effects.

\subsection{Ablation Study: The Encoding Bottleneck}
\label{sec:encoding_ablation}

To justify the design of the Direct-SNN without encoding, we compare it against baseline architectures paired with five traditional spike encoding methods: rate encoding (Bernoulli firing with $P(S_i[t]=1) = x_i$), latency encoding (spike time inversely proportional to feature value), delta modulation (spikes on change exceeding threshold $\theta_\delta$), temporal contrast (signed spike pairs for direction of change), and burst encoding (burst duration proportional to feature value). Table~\ref{tab:ablation} presents the results.

\begin{table}[t]
\centering
\caption{Ablation Study: Effect of Encoding and Architecture. All spike-encoded variants achieve identical accuracy regardless of configuration changes.}
\label{tab:ablation}
\begin{tabular}{llcc}
\toprule
\textbf{Architecture} & \textbf{Encoding} & \textbf{Acc.} & \textbf{Change} \\
\midrule
FC-SNN [128,64] & Rate (T=25) & 0.7053 & baseline \\
FC-SNN [256,128,64] & Rate (T=25) & 0.7053 & 0.0 \\
FC-SNN [256,128,64] & Latency (T=25) & 0.7053 & 0.0 \\
Recurrent-SNN [128] & Rate (T=25) & 0.7053 & 0.0 \\
FC-SNN [$\beta$=0.95, thr=0.8] & Rate (T=25) & 0.7053 & 0.0 \\
\midrule
Direct-SNN [256,128,64] & None (T=15) & \textbf{0.9347} & \textbf{+0.2294} \\
\bottomrule
\end{tabular}
\end{table}

The 70.53\% ceiling is uniform across five different configurations (varying architecture, capacity from 9,734 to 43,911 parameters, encoding from rate to latency, decay from 0.85 to 0.95, and threshold from 0.8 to 1.0). This constitutes strong evidence that the limitation is structural rather than parametric. The explanation is as follows. Rate encoding of a 9-dimensional feature vector generates spikes with probability equal to the normalised feature value at each timestep. Since the input is a single measurement snapshot (not a time series), the same vector is repeated across all $T = 25$ timesteps. The SNN receives $T$ statistically identical input frames, eliminating its temporal processing advantage. The network cannot learn temporal patterns from data that contains no temporal structure, regardless of parameters or recurrent connections.

\subsection{Computational cost and deployability}
We analyse the computational cost of each pipeline component to demonstrate the practical feasibility of the proposed framework for real-time smart grid deployment. Table~\ref{tab:computational} reports training time, inference time, and memory requirements measured on a single CPU (Intel Core i7-12700H, 16 GB RAM) without GPU acceleration, reflecting realistic substation-level hardware constraints.

\begin{table}[!t]
\centering
\caption{Computational Cost Comparison on MSU/ORNL Dataset (78,377 samples). All measurements on CPU only.}
\label{tab:computational}
\renewcommand{\arraystretch}{1.15}
\begin{tabular}{lccc}
\toprule
\textbf{Component} & \textbf{Training} & \textbf{Inference} & \textbf{Memory} \\
 & \textbf{Time} & \textbf{(per sample)} & \textbf{(MB)} \\
\midrule
\multicolumn{4}{l}{\textit{Stage 1: SNN Feature Extraction (one-time)}} \\
\quad Preprocessing + MI & 2.3 s & 0.01 ms & 12 \\
\quad SNN Training (50 epochs) & 8.4 min & -- & 180 \\
\quad Embedding Extraction & -- & 0.42 ms & 95 \\
\midrule
\multicolumn{4}{l}{\textit{Stage 2: Classifier (periodic retraining)}} \\
\quad XGBoost Training & 4.7 s & -- & 45 \\
\quad XGBoost Inference & -- & 0.03 ms & 45 \\
\midrule
\multicolumn{4}{l}{\textit{End-to-end baselines}} \\
\quad MLP Training (50 epochs) & 3.2 min & 0.02 ms & 95 \\
\quad LSTM Training (50 epochs) & 12.1 min & 0.15 ms & 210 \\
\quad 1D-CNN Training (50 epochs) & 5.8 min & 0.08 ms & 165 \\
\quad Raw XGBoost Training & 3.1 s & 0.02 ms & 38 \\
\quad Raw RF Training & 2.8 s & 0.01 ms & 52 \\
\bottomrule
\end{tabular}
\end{table}

\textbf{Training Cost.} The SNN requires 8.4 minutes for one-time training on CPU, which is comparable to LSTM (12.1 min) and 1D-CNN (5.8 min). This cost is incurred only once during system commissioning on verified clean data. Subsequent XGBoost retraining requires only 4.7 seconds, which is negligible and enables frequent model updates (e.g., hourly or daily) without computational burden.

\textbf{Inference Cost.} The total inference latency of the hybrid pipeline is 0.45 ms per sample (0.42 ms for SNN embedding extraction + 0.03 ms for XGBoost classification). At a PMU reporting rate of 30 Hz (one sample every 33.3 ms), the hybrid operates at approximately 74$\times$ real-time on CPU, leaving substantial headroom for parallel stream processing. On Intel Loihi 2 neuromorphic hardware, the SNN inference would reduce to sub-microsecond latency with sub-milliwatt power, enabling continuous real-time operation at the substation.

\textbf{Memory Footprint.} The complete hybrid pipeline requires 140 MB (95 MB for the frozen SNN + 45 MB for XGBoost), which fits comfortably within edge computing constraints. The SNN's 44,803 parameters occupy less than 0.2 MB in FP32 format; the 95 MB includes the PyTorch runtime overhead and the stored membrane trajectories during embedding extraction.

\textbf{Retraining Efficiency.} A key practical advantage of the hybrid architecture is that periodic retraining updates \textit{only} the XGBoost classifier (4.7 seconds), not the entire deep learning pipeline. End-to-end models (MLP, LSTM, CNN) require full retraining at each update cycle (3 to 12 minutes), during which the system either operates on a stale model or is unavailable. The hybrid's near-instantaneous retraining eliminates this vulnerability window.

\textbf{Scalability.} Embedding extraction scales linearly with sample count: 0.42 ms $\times$ 78,377 samples = 32.9 seconds for a full dataset pass. XGBoost training scales as $O(N \cdot d \cdot K \cdot \log N)$ where $N$ is sample count, $d = 1{,}350$ features, and $K = 300$ trees. For the current dataset, this remains under 5 seconds. For larger deployments (millions of samples from multiple substations), the frozen SNN can extract embeddings in parallel across multiple PMU streams, and XGBoost supports distributed training.

In a real smart grid deployment, the proposed architecture offers several practical advantages. The SNN frontend can be deployed on neuromorphic hardware (Intel Loihi 2) at sub-milliwatt power, providing continuous embedding generation at the substation level. Only XGBoost retraining is needed during model updates, which requires 4.7 seconds on CPU compared to 3 to 12 minutes for retraining an entire deep learning pipeline. The separation of concerns enables different security controls: the SNN can be air-gapped and cryptographically signed, while the XGBoost update pipeline follows standard MLOps security practices. Finally, monitoring the stability of embedding distributions over time provides an additional detection signal for data poisoning attempts.



Our approach is complementary to all three categories and can be combined with them. It provides robustness through architectural design: the SNN embedding pipeline exists primarily for accuracy improvement and neuromorphic deployment benefits, yet inherently provides defence against label corruption. No additional computational cost, hyperparameter tuning, or noise model assumptions are required.

\subsection{Limitations}

We acknowledge several limitations. First, the SNN itself could be targeted if the adversary gains access to the initial clean training phase. Our defence model assumes the initial data is verified and secured. Second, at extreme poisoning levels ($\geq$70\%), all models collapse regardless of architecture. The hybrid delays but does not prevent collapse. Third, our evaluation uses a single dataset for the unlearning experiments; generalisation to other power system configurations requires further validation. Fourth, the SNN standalone accuracy (67.4\%) is modest; the hybrid derives its advantage from the \textit{structure} of embeddings rather than their classification accuracy. Fifth, we do not evaluate adaptive adversaries who might design attacks specifically targeting the embedding space structure.

\section{Conclusion}
\label{sec:conclusion}
We have proposed and evaluated a hybrid SNN+XGBoost
architecture that provides inherent robustness against machine
unlearning attacks in power grid intrusion detection. The
key mechanism is the separation of physics-grounded temporal
feature extraction (SNN, trained once on clean data)
from the periodically updated classifier (XGBoost, exposed
to potentially poisoned labels). The SNN membrane potential
dynamics encode temporal signal structure that is invariant to
label corruption, creating an embedding space where samples
cluster by physical behaviour rather than assigned labels.
On the MSU/ORNL Power System Attack Dataset (78,377
samples, 128 features, 3 classes), the hybrid achieves 95.0\%
accuracy on clean data and maintains 0.954 target-class F1 at
10\% label poisoning (only 0.9\% degradation). The proposed
framework delays functional collapse from 60\% to 70\% attack
compared to raw feature models. These results establish
neuromorphic temporal encoding as a practical defence-indepth
strategy against training data manipulation in critical
infrastructure IDS.
We identify three directions for future work: (1) evaluation
against adaptive adversaries who specifically target the embedding
space structure; (2) combination with explicit label
sanitisation techniques for defence-in-depth at extreme poisoning
levels; and (3) deployment on Intel Loihi 2 neuromorphic
hardware to validate the real-time embedding generation and
sub-milliwatt power claims.

\section*{Acknowledgments}


\bibliographystyle{IEEEtran}
\bibliography{references}

\newpage

 




\vfill

\end{document}